\newif\ifextended
\extendedtrue
\documentclass{article}
\usepackage{spconf,amsmath,graphicx}
\usepackage[utf8]{inputenc} % allow utf-8 input
\usepackage{amsfonts}       % blackboard math symbols
\usepackage{tikz}

\newenvironment{proof}{\textbf{Proof:}}{\hfill$\square$}

\newcommand{\N}{\mathbb{N}}

\newtheorem{definition}{Definition}[section]
\newtheorem{theorem}[definition]{Theorem}

\newtheorem{proposition}[definition]{Proposition}
\newtheorem{rem}[definition]{Remark}

\renewcommand{\P}{{\mathbb{P}}}
\newcommand{\E}{{\mathbb{E}}}
\newcommand{\R}{{\mathbb{R}}}

\newcommand{\argmin}{\operatorname*{argmin}}

\newcommand{\Ccal}{\mathcal{C}}
\renewcommand{\epsilon}{\varepsilon}
\renewcommand{\phi}{\varphi}

\title{Improving Accuracy of Nonparametric Transfer Learning via\\Vector Segmentation}
\name{Vincent Gripon \and Ghouthi B. Hacene \and Matthias L\"owe \and Franck Vermet}
\threeauthors
  {Vincent Gripon, Ghouthi B. Hacene}
  {IMT Atlantique\\
    Electronics Dept.}
  {Matthias L\"owe}
  {University of M\"unster\\
    Fachbereich Mathematik\\und Informatik
  }
  {Franck Vermet}
  {Université de Brest\\
    Laboratoire de Math\'ematiques
  }

\newcommand{\vep}{\varepsilon}

\begin{document}

\maketitle

\begin{abstract}
  Transfer learning using deep neural networks as feature extractors has become increasingly popular over the past few years. It allows to obtain state-of-the-art accuracy on datasets too small to train a deep neural network on its own, and it provides cutting edge descriptors that, combined with nonparametric learning methods, allow rapid and flexible deployment of performing solutions in computationally restricted settings. In this paper, we are interested in showing that the features extracted using deep neural networks have specific properties which can be used to improve accuracy of downstream nonparametric learning methods. Namely, we demonstrate that for some distributions where information is embedded in a few coordinates, segmenting feature vectors can lead to better accuracy. We show how this model can be applied to real datasets by performing experiments using three mainstream deep neural network feature extractors and four databases, in vision and audio.
\end{abstract}

\section{Introduction}

Transfer learning consists in training a learning method on a first dataset to be used on a second, distinct one. In this context, using Deep Neural Networks (DNNs)~\cite{pan2010survey,yosinski2014transferable,girshick2014rich} (in particular convolutional neural networks on vision datasets) has become increasingly popular over the past few years. Indeed, the features extracted by state-of-the-art deep neural networks are so good that they allow, in some settings, to reach the best known accuracy when applied to other datasets and combined with simple classification routines.
One of the key interest in using transfer learning methods is to avoid the heavy computational cost of training DNNs. Therefore, it is possible to exploit their accuracy on embedded devices such as small robots or smartphones~\cite{qiu2016going,iandola2016squeezenet}. In this context, nonparametric methods such as $k$-Nearest Neighbors ($k$-NN) are particularly attractive for their ability to handle both class-incremental and example-incremental properties~\cite{polikar2000learn++,learn++}.

It is interesting to point out that DNNs are trained to extract features well suited to perform a given classification task. In order for these features to become usable in other contexts (e.g. a new classification task), broad databases containing a large variety of classes should be used. As a direct consequence, it is expected that a significant part of the extracted feature vectors is useless for solving the new task at hand.
Consequently, e..g., in the field of approximate nearest neighbor search for classification it is often observed that methods based on Product Quantization~\cite{jegou2011product} and its derivatives can lead for certain choices of parameters to better performance than an exhaustive search on raw data. In these methods, the search space is split into quantized subspaces in which the search is performed independently.

In this paper we are interested in showing that, more generally, segmentation of feature vectors (obtained with pretrained DNNs) in multiple subvectors for which the search is performed independently can result in higher overall accuracy. We are interested in answering the following questions:
\begin{itemize}
\item Are there simple convincing mathematical models in which such improvement exists?
\item How do such improvements depend on parameters?
\item Does this apply to real world data?
\end{itemize}

To answer these questions, we describe the mathematical core of the classification procedure in Section 2. Section 3 contains an example where this procedure is not successful as well as a couple of situations where segmentation not only helps, but does provide the right class of the test pattern with probability converging to 1 as dimension becomes large, while a comparison of the Euclidean distance as well as a comparison coordinate by coordinate fails with probability at least one half. Section 4 contains experiments on real datasets. Section 5 is a conclusion.

\section{Classification by segmentation}
In this section we will give a mathematical framework of the procedure we have in mind. We start with classes of data $\Ccal_1, \ldots, \Ccal_K \subseteq \R^d$, where we always assume that the dimension $d$ is a large parameter. To simplify matters assume that all of these classes have $M$ elements.
Now choose $c$, such that $c$ divides d and write $\R^d = \bigotimes_{j=1}^c \R^{d/c}$. The $j'th$ of these subspaces will also be denoted by $(\R^{d/c})_j$. For each $j$ take a dictionary $D_j$ of $nK$ segments such that $D_j$ contains $n$ segments of each class $\Ccal_k$ drawn uniformly at random. Thus, we write each word $w \in \Ccal_k$, $k=1, \ldots, K$ as $w=w_1 \circ \ldots \circ w_c$, where each of the $w_j \in (\R^{d/c})_j$ and ``$\circ$'' denotes concatenation. For each class $\Ccal_k$, $k=1, \ldots, K$ we pick $n$ segments of words $w_j$ in $\Ccal_k$ uniformly at random without replacement and put them into the dictionary $D_j$.
Given a fresh word $z$ we parse it in the same way into $z=z_1 \circ \ldots \circ z_c$. Then for each $j= 1, \ldots, c$ we find
\begin{equation}\label{distance}
\tilde w_j := \argmin\{||w_j - z_j||_{d/c} : w_j \in D_j\}.
\end{equation}
Here $||\cdot||_{d/c}$ denotes Euclidean distance in $\R^{d/c}$.
Let the random variable $U_j$ take the value $k$, if $\tilde w_j$ is the $j$'th segment of a word $w \in \Ccal_k$. If several words minimize the distance in \eqref{distance} we let $U_j$ take any of the values of the classes corresponding to these words with equal probability, as a tie-breaking rule.
Finally one takes
$$
\chi(z) := \mathrm{argmax}_k \sum_{j=1}^c \{\mathbb{I}_{U_j=k}, k=1, \ldots K\}
$$
hence the class that is most often found by the above procedure. Again we add a tie-breaking rule, if this class is not unique.
The $c$-segmentation procedure assigns $z$ the class $\chi(z)$.

\section{Situations where segmentation is or is not favorable}
%% In this section we will try to find out conditions that favor the use of the $c$-segmentation procedure described in the previous section.
\subsection{When segmentation does not help}
We start with an example that shows that the data need to have a special structure for the segmentation technique to be supportive. In this subsection we will assume that we only have two classes $\Ccal_1$ and $\Ccal_2$. These are built in the following way. Take $m_1, m_2 \in \{-1,+1\}^d$ uniformly at random as ``base vectors'' of the two classes. Assume $0<\varrho< \frac 12$. Then for $\mu=1, \ldots, M$ let $Y^\mu, \tilde Y^\mu \in \{-1,+1\}^d$ be i.i.d. vectors with i.i.d. coordinates such that
$\P(Y^1_1=1)=1-\varrho \ \mbox{and}\ \P(Y^1_1=-1)=\varrho.$
We define
$
\Ccal_1:=\{Y^\mu \times m_1, \mu =1, \ldots, M\}
\ \mbox{and}\ \
\Ccal_2:=\{\tilde Y^\mu \times m_2, \mu =1, \ldots, M\}.
$
Here the multiplication is pointwise, i.e. $(Y^\mu \times m_1)_i:= Y^\mu_i m_{1,i}$. Also let us assume that for $c \in \N$ such that $c$ divides $d$ and each $1 \le j \le c$ we take a dictionary $D_j$ consisting of two segments, one, $w_j^1$, belonging to class $\Ccal_1$, one, $w_j^2$, belonging to class $\Ccal_2$, only. This will help to facilitate computations. In this setting we claim that $c$-segmentation does not improve the accuracy of the naive Euclidean approach.
\begin{proposition}
In the above situation assume that
%$1 \ll c \ll d$, that
 $c$ and $d/c$ are odd (to avoid the discussion of tie-breaks) and that $w$  is distributed like a word from $\Ccal_1$ (but independent of all words in all classes). Then there is a number $I>0$ such
$$
\lim_{d \to \infty} \frac 1d \log \P(w \text{ is classified $\Ccal_2$ with } c=1) = -I.
$$
Whilst
$$
\lim_{d \to \infty} \frac 1d\log \P(w \mbox{ is classified } \Ccal_2\ \mbox{with } 1 \ll c \ll d) = -I/2.
$$
\end{proposition}
\ifextended
\begin{proof}
First of all notice that, if $w$ and any of the other words, say $w^1$, differ in a coordinate $i$, they do so by 2. Hence we have that
$||w-w^1||= \sqrt 2 d_H(w,w^1)$
where $d_H$ denotes Hamming distance and this is also true for any subspace of $(\R^{d/c})\subseteq \R^d$. We will first compute the probability that the $j$'th segment $w_j$ of $w$ is misclassified:
%\begin{eqnarray*}
$\P(d_H(w_j, w_j^1) \ge d_H(w_j, w_j^2)) = \P(\sum_{i=1}^{d/c} X_i \ge 0)$.
  Here $X_i$ is a random variable such that
\begin{eqnarray*}
X_i = \begin{cases}
1 & \mbox{if  } |w_{j,i}-w_{j,i}^1| > |w_{j,i}-w_{j,i}^2|\\
0 & \mbox{if  } |w_{j,i}-w_{j,i}^1| = |w_{j,i}-w_{j,i}^2|\\
-1 & \mbox{if  } |w_{j,i}-w_{j,i}^1| < |w_{j,i}-w_{j,i}^2|
\end{cases}
\end{eqnarray*}
A quick calculation shows that
\begin{eqnarray*}
\P(X_i=v) = \begin{cases}
\varrho(1-\varrho) & \mbox{if  } v=1\\
\frac 1 2 & \mbox{if  } v=0\\
\frac 12 (1- 2\varrho+2 \varrho^2) & \mbox{if  }  v=-1.
\end{cases}
\end{eqnarray*}
In particular $\mu:=\E X_i= 2\varrho (1-\varrho)-\frac 12 <0$.
Thus by Cram\'er's theorem (cf. \cite{dembozeitouni}, Theorem 2.1.24) and the convexity of the rate function there, we get
\begin{eqnarray*}\label{calc_misclassi}
\lim_{d/c \to \infty}\frac c d \log \P(d_H(w_j, w_j^1) \ge d_H(w_j, w_j^2))&=&\\ \lim_{d/c \to \infty}\frac c d \log
\P(\sum_{i=1}^{d/c} X_i \ge 0)&=&-I(0)
\end{eqnarray*}
where
$
I(x) = \sup_{t\in \R} \left\{tx - \log E(e^{tX_1})\right\}
$
and in particular $I(0)>0$. This means that the probability of a misclassification in a segment of length $d/c$ asymptotically behaves like $e^{-\frac dc  I(0)}$.
The probability to completely misclassify $w$ is now given by
$
\P(w \mbox{ misclassified})= \P(\sum_{j=1}^c Y_j \ge c/2)
$
Here $Y_j$ is the indicator for the event that $w_j$ is closer to $w_j^2$ than to $w_j^1$. Again we will apply Cram\'ers theorem to compute the asymptotics of this probability. Recall the fact that $Y_j$ are i.i.d. Bernoullis with success-probability $p \sim e^{-\frac dc  I(0)}$ and that the rate function in the large deviations principle for Bernoullis with success-probability $p$ is the relative entropy (this is actually a special case of Theorem 2.1.10 in \cite{dembozeitouni})
$
H(x|p) = x \log \frac x p + (1-x) \log \frac{1-x}{1-p}.
$
For our choice $p$ we obtain especially
$$
H(\frac 12 |p) = -\log 2  - \frac 12 \log p - \frac 12 \log(1-p) \sim -\log 2 +\frac d {2c} I(0).
$$
Thus
\begin{eqnarray*}
&&\lim \frac 1 d  \log \P(w \mbox{ misclassified})\\ &=& \lim \frac c d \frac 1 c \log \P(\sum_{j=1}^c Y_j \ge c/2)\\
&\sim & \lim \frac c d (\log 2 -\frac d {2c} I(0))=- \frac 12 I(0).
%\\&=&- \frac 12 I(0).
\end{eqnarray*}
On the other hand, using \eqref{calc_misclassi} for $c=1$, i.e. using just the Euclidean distance in $\R^d$ we obtain
$$
\lim_{d \to \infty}\frac 1 d \log \P(d_H(w, w^1) \ge d_H(w, w^2))=-I(0),
$$
hence the rate  function of the probability of a misclassification with $c=1$  is twice  the rate function of the above probability.
\end{proof}
\else
\begin{proof}
  Refer to the extended version~\cite{extended}.
\end{proof}
\fi

\ifextended
\begin{rem}
It is also interesting to compare the accuracy of the $c$-segmentation procedure to the other ``natural'' classification technique, the coordinate-by-coordinate comparison. Here a misclassification occurs, if $\sum_{i=1}^d \tilde X_i \ge 0$ and the random variables $\tilde X_i$ take two values only:
$$
\P(\tilde X_i=v) = \begin{cases}
\varrho(1-\varrho)+ \frac 14  & \mbox{if  } v=1\\
\frac 12 (1- 2\varrho+2 \varrho^2) + \frac 14 & \mbox{if  }  v=-1
\end{cases}
$$
Indeed the summand $\frac 14$ in each of the lines reflects the tie-breaking rule, if two coordinates agree.
Again by Cram\'er's theorem
$
\lim_{d \to \infty} \frac 1 d \log \P( \sum_{i=1}^d \tilde X_i \ge 0) = - \tilde I(0)
$
with
$
\tilde I(x)= \sup_{t\in \R} \{tx - \log E(e^{t \tilde X_1})\}.
$
Now with the notation from the previous proof
$$
I(0)=-\inf_{t} \log \E e^{t X_1} \quad \mbox{and } \, \tilde I(0)=-\inf_{t} \log \E e^{t \tilde X_1}.
$$
But for all $t \in \R$
\begin{eqnarray*}
\E e^{t X_1}&=& \varrho (1-\varrho) e^{t}+ \frac 12 +  \frac 12 (1- 2\varrho+2 \varrho^2) e^{-t}\\
&\le& \varrho (1-\varrho) e^{t}+\frac 12 \cosh(t) +  \frac 12 (1- 2\varrho+2 \varrho^2) e^{-t}\\&=&\E e^{t \tilde X_1}.
%\\&=&\E e^{t \tilde X_1}
\end{eqnarray*}
Therefore $I(0) \ge \tilde I(0)$
such that in this case the decision based on the Euclidean $c=1$ norm is the best of the proposed segmentation methods.
%% \item

It is also worth mentioning, that although in this case it seems not useful to segment at all or into coordinates, segmentation with $1 \ll c\ll d$ pieces still yields a  classification with a probability close to 1.
%% \end{itemize}
\end{rem}
\fi

\subsection{When segmentation does help}
In the previous paragraph we saw that there are natural situations where the simplest case, when one does not partition vectors at all, is the best. However, the situation described there is close to a situation where using pieces of size $1 \ll c\ll d$ not only gives a better result than $c=1$ and $c=d$, but also the results for the latter choice are useless.

We will start by describing a basic situation and then discuss possible extensions. All these models are influenced by the observation that the data classified in \cite{gripon_classification} seem to suffer from occasional large outliers.  %% For illustration, we consider the feature vectors obtained by proceeding the images of CIFAR10 through the Inception V3~\cite{szegedy2015rethinking} DNN, and we plot the histograms of values of the coordinates of two vectors taken in the same class or in two different classes in Figure~\ref{figure1}.

 Our first basic situation will be given by $K=2$ classes $\Ccal_1$ and $\Ccal_2$
where
\begin{eqnarray*}
&&\Ccal_1:=\{Y^\mu + m_1, \mu =1, \ldots, M\} \\&\text{and} &\Ccal_2:=\{\tilde Y^\mu + m_2, \mu =1, \ldots, M\}.
\end{eqnarray*}
%and
%$$
%\Ccal_2:=\{\tilde Y^\mu + m_2, \mu =1, \ldots, M\}.
%$$
This time (for the sake of keeping things easy), $m_1= (0, \ldots, 0)$ and
$
m_2=(1,\underbrace{0, \ldots,0}_{l-1 \text{ times}}, 1,  \underbrace{0, \ldots,0}_{l-1 \text{ times}},1 \ldots 0)
$
i.e. $m_2$ has a $1$ at regular positions. Moreover the $Y^\mu$, and $\tilde Y^\mu$ are i.i.d vectors in $\R^d$ such that
$\P(Y^1_1=N)=\P(\tilde Y^1_1=N)=p=1-\P(Y^1_1=0)=1-\P(\tilde Y^1_1=0)$
and $p$ and $N$ will be chosen in the sequel.

Given $c$ for each segment $1 \le j \le c$ we will again take a short dictionary $D_j$ consisting of one segment $w_j^1$ of a word from $\Ccal_1$ and one segment $w_j^2$ from a word of $\Ccal_2$.
Assume again we want to classify a word $w$ that is distributed like a word from $\Ccal_1$ (but independent of all words in all classes).
We start with the observation, that for small $l$ and small $p$ the coordinate by coordinate comparison will fail.
\begin{proposition}\label{prop_pixel}
Assume that $l \le d^{1/4}$, $N>0$, and that $p \le \frac 1l$. Then for $c=d$, i.e. the coordinate by coordinate comparison,
\begin{equation}\label{wrong1}
\P(w \text{ is classified correctly} ) \to \frac 12 \quad \text{as $d\to \infty$}.
\end{equation}
\end{proposition}
\begin{rem} Observe that the situation described in \eqref{wrong1} is a worst case scenario when one has two classes only. Indeed if the probability on the right were even smaller than $\frac 12$ one could use the reverse method and decide just the opposite of the proposed classification to get a better result.
\end{rem}
\begin{proof}
Due to the independence of the random parts of the coordinates of the words, we may assume without loss of generality that all the segments from class one in $D_j$ stem from the same word $w^1= m_1+ Y$, all the segments from class two stem from the same word $w^2= m_2+ \tilde Y$. Moreover, we write $z=m_1 + \overline{Y}$.
Consider the set
$$
S:=\{i: \exists n, i=1+nl \text{ or } Y_i=N \text{ or } \tilde Y_i=N \text{ or } \overline{Y_i}=N \}
$$
According to our assumptions, $|S| \le d^{1/3}$ with high probability, i.e. with probability converging to 1 as $d \to \infty$ and thus
$
|\{1, \ldots,d\}\setminus S| \ge d - d^{1/3}
$
with high probability. But for all coordinates $i$ in $\{1, \ldots,d\}\setminus S$ one has
that $|w_i-w_i^1|=|w_i-w_i^2|$, therefore the tie-breaking rule decides with probability one half for class $\Ccal_1$. As the fluctuations of this random decision by the Central Limit Theorem are of order $\sqrt d$ and therefore larger than any ``signal'' one might obtain from $S$, the statement follows.
\end{proof}

But also the Euclidean distance $c=1$ fails as a classification rule for a wide range of parameters.
\begin{proposition}\label{prop_euclid}
If  $l \to\infty$ and $p \gg \max(\frac 1d,  \frac 1{N^2l})$, we have for the Euclidean distance rule $c=1$
\begin{equation}\label{wrong2}
\P(w \text{ is classified correctly} ) \to \frac 12 \quad \text{as $d\to \infty$}.
\end{equation}
\end{proposition}
\ifextended
\begin{proof}
Again suppose that all the two words in the  dictionary are $w^1= m_1+ Y\in \Ccal_1$ and $w^2= m_2+ \tilde Y \in \Ccal_2$. Moreover, we write $w=m_1 + \overline{Y}$.
Observe that
$$
||w-w^1||^2= N^2 \sum_{i=1}^d \mathbb{I}_{Y_i \neq \overline{Y_i}}
$$
and
$$||w-w^2||^2= N^2 \sum_{i=1 \atop i\!\! \! \!\! \mod\! l\ne 1}^d \mathbb{I}_{\overline{Y_i} \neq \tilde Y_i}+\sum_{i=1\atop i\!\! \!\!\!  \mod\! l= 1}^d (1+ \tilde{Y_i}-\overline{Y_i})^2.
$$
With large probability, the term $||w-w^1||^2$ is of order $N^2 dp$ and
the last sum in $||w-w^2||^2$ is of order $\frac dl (1+2 N^2p)$, which implies that this last term will be negligible with respect to $||w-w^1||^2$,  if we choose $p\gg\frac 1{N^2l}$ and $l \to\infty$.
%and
%$$
%||w-w^2||^2= N^2 \sum_{i=1}^d \mathbb{I}_{\overline{Y_i} \neq \tilde Y_i}+l
%$$

Moreover, for $p\gg \frac 1d$ both sums $\sum_{i=1}^d \mathbb{I}_{Y_i \neq \overline{Y_i}}$ as well as
$\sum_{i=1}^d \mathbb{I}_{\overline{Y_i} \neq \tilde Y_i}$ obey a Central Limit Theorem with the same parameters (we can omit the condition $i\!\!\!\mod l\ne 1$, since $\frac dl\ll d$). Therefore,
$$
\P\left(\sum_{i=1}^d \mathbb{I}_{\overline{Y_i} \neq \tilde Y_i}<\sum_{i=1}^d \mathbb{I}_{Y_i \neq \overline{Y_i}}\right)
\to \frac 12,
$$
which gives the result.
\end{proof}

\else
\begin{proof}
  Refer to the extended version~\cite{extended}.
\end{proof}
\fi

The question remains, of course, whether there is any segmentation method that works in this case. Fortunately, the answer is yes.
\begin{proposition}\label{prop_segment}
If $l\ll \frac 1p$ the $c$-segmentation rule with $c=d/l$ works, more precisely
$$
\P(w \text{ is misclassified}) \to 0, \quad \mbox{ as $l \to \infty$ and $d \to \infty$}.
$$
\end{proposition}
\ifextended
\begin{proof}
With the notation of the previous two proofs, let $\chi_j$ be the indicator for the event that the $j'th$ block of $w$, $w_j$, is classified correctly. With $c=d/l$ the segments have length $l$ and therefore the base vectors $m_1$ and $m_2$ of $\Ccal_1$ and $\Ccal_2$, respectively, have Euclidean distance 1 in each segment. Thus $\chi_j=1$, if
\begin{eqnarray*}
Y_i=\tilde Y_i= \overline{Y_i}=0 \text{ for all } i=\frac{(j-1)d}{l+1}, \frac{(j-1)d}{l+2}, \ldots, \frac{jd}{l-1}.
\end{eqnarray*}
Since we assume that $p \ll \frac 1l$, we have that for any given $\vep >0$ and any fixed $j$
\begin{eqnarray*}
\P(Y_i=\tilde Y_i= \overline{Y_i}=0, \forall i=\frac{(j-1)d}{l+1}, \frac{(j-1)d}{l+2}, \ldots, \frac{jd}{l-1}) \\
= (1-p)^{3l} \sim e^{-3pl} \ge 1- \vep,
\end{eqnarray*}
as $l \to \infty$.
Moreover these events are i.i.d. for different $j$. Therefore with high probability the majority of the $\chi_j$ will be 1. This proves the proposition.
\end{proof}
\else
\begin{proof}
  Refer to the extended version~\cite{extended}.
\end{proof}
\fi

Altogether we have shown
\begin{theorem}\label{thm1}
In the model described above assume that $d,l\to \infty$, $l \le d^{\frac 14}$, $p \ll \frac 1 l$, but $p \gg \max(\frac 1d,  \frac 1{N^2l})$,  then for $c=d$ and $c=1$ we have
$
\P(w \text{ is misclassified}) \to 1/2,
$
while for $c= d/l$,  we have
$
\P(w \text{ is misclassified}) \to 0.
$
\end{theorem}

Up to now we have just discussed the basic example to illustrate which statistical properties of the classes favor segmentation in the classification process. Let us comment on some variants of the above model. A first natural extension of the model is to consider more than two classes. In the above setting it is obvious that we can build up to $l+1$ classes, where again
$
\Ccal_1:=\{Y^{1,\mu}+m_1, \mu =1, \ldots, M\}
$
with $m_1=0$, and for $k=2, \ldots, K$,
$
\Ccal_k:=\{Y^{k,\mu} + m_k, \mu =1, \ldots, M\},
$
where all the $(Y^{k,\mu})$ are i.i.d. random vectors in $\R^d$ with
$\P(Y^{k,\mu}_i=N)=p=1-\P(Y^{k,\mu}_i=0)$
and the vectors $(m_k)_{k=2,\ldots,K}$ are concatenations of strings of length $l$, such that for each of these strings contains
all coordinates but one are 0, the remaining coordinate is 1, and the $1$s are placed at different positions for different $m_k$. Then Theorem \ref{thm1} translates to
\begin{theorem}\label{thm2}
In the model described above assume that $d,l\to \infty$, $l \le d^{\frac 14}$, $p \ll \frac 1 l$, but $p \gg \max(\frac 1d,  \frac 1{N^2l})$, then for $c=d$ and $c=1$ we have
$
\P(w \text{ is misclassified}) \to 1/K
$
while for $c= d/l$,  we have
$
\P(w \text{ is misclassified}) \to 0.
$
\end{theorem}
\ifextended
\begin{proposition}\label{prop_pixel2}
Assume that $l \le d^{1/4}$, $N>0$, and that $p \le \frac 1l$. Then for $c=d$, i.e. the coordinate by coordinate comparison
\begin{equation}\label{wrong}
\P(w \text{ is classified correctly} ) \to \frac 1K \quad \text{as $d\to \infty$}.
\end{equation}
\end{proposition}
The proof can be copied almost literally from the proof of Proposition \ref{prop_pixel}. Similarly Proposition \ref{prop_euclid} can be translated to
\begin{proposition}\label{prop_euclid2}
If $\sqrt N >l$ and $p \gg \frac 1d $ we have for the Euclidean distance rule $c=1$
\begin{equation}\label{wrong2}
\P(w \text{ is classified correctly} ) \to \frac 1K \quad \text{as $d\to \infty$}.
\end{equation}
\end{proposition}
Of course, it is not difficult to check that the proofs given for the case of two classes translate to the case of several classes. On the other hand, it is also evident, that correct classification can only become difficult, if more classes are available. The crucial question is thus, if the $c$-segmentation rule also gives the right decision with more than two segments. However, checking the proof of Proposition \ref{prop_segment} it is clear that also for more than two classes we have
\begin{proposition}\label{prop_segment2}
If $l\ll \frac 1p$ the $c$-segmentation rule with $c=d/l$ works, i.e. more precisely
$$
\P(w \text{ is misclassified}) \to 0.
$$
as $d \to \infty$.
\end{proposition}
So altogether the case of two classes is generic. Therefore we will discuss other variants of the model for this case only.
\else
Altogether the case of two classes is generic (\ref{extended}). Therefore we will discuss other variants of the model for this case only.
\fi

Another obvious modification of the model at the beginning of this subsection one might discuss is the influence of a larger dictionary. So let us assume now that the dictionaries $D_j$ contain $\nu$ segments of each class, i.e.
$
D_j:=\{w^{1,1}_j, \ldots w^{1,\nu}_j, w^{2,1}_j, \ldots w^{2,\nu}_j\}
$
and again the words are of the form $w^{1,\mu}:= m_1+ Y^{1,\mu}$, $w^{2,\mu}:= m_2+ Y^{2,\mu}$.
Again we will check whether Propositions \ref{prop_pixel} to \ref{prop_segment} remain true.
For the proof of Proposition \ref{prop_pixel} we let $w= m_1+Y$ be distributed as a word from class one. Define the set $S$ in the proof of Proposition \ref{prop_pixel} now as the set of coordinates that are not of the form $1+jl$ and such that none of the Bernoulli's in the dictionaries is 1. Then again $|S|\le d^{1/3}$ with high probability, which implies that Proposition \ref{prop_pixel} remains true.
However, the behaviour of the Euclidean distance rule Proposition \ref{prop_euclid} improves, if the dictionaries become larger. Indeed $w$ is classified correctly, if there exists a word $w^{1,n} \in \{w^{1,1}, \ldots, w^{1,\nu}\} $ such that
$$
\sum_{i} \mathbb{I}_{Y_i^{1,n} \neq Y_i} <  \sum_{i} \mathbb{I}_{ Y_i^{2,s} \neq Y_i}
\quad \forall s=1, \ldots, \nu.
$$
The probability that this holds true for a fixed $n$ is asymptotically $(\frac 12)^\nu$. So the probability that such an $n$ does not exist is given by $1-(1-(\frac 12)^\nu)^\nu$, which is smaller than $1/2$ but for $\nu$ not depending on $d$ still not $0$. However, also the accuracy of the $c$-segmentation method with $c=d/l$ as in Proposition \ref{prop_segment} improves and this basically for the same reasons: If there is one segment of a word of class $\Ccal_1$ in the j'th dictionary such that all its $Y$-variables in this segment are 0, one classifies $w$ correctly. And this probability, of course, increases, as $\nu$ becomes larger.

One might, of course, ask which features of the model discussed in this subsection are decisive for the $c$-segmentation method to be favorable. These features are:
\begin{itemize}
\item[a)] The vectors in each class are {\sl rare} but {\sl large} perturbations of a base vector. Most of the coordinates of the base vectors of two distinct classes agree.
\item[b)] The perturbations are much rarer than the frequencies of the coordinates in which the base vectors disagree.
\end{itemize}
%% \ifextended
%% Property a) above leads to classes that are random perturbations of a given base word or vector.
%% However, sometimes one might want to model situations where the properties of a given class are more of a statistic nature, i.e. where the elements of one class only share a common distribution.
%% This can easily be done by considering e.g.
%% $$
%% \Ccal_1:=\{Y^{1,\mu} + m_1, \mu =1, \ldots, M\}
%% $$
%% and
%% $$
%% \Ccal_2:=\{Y^{2,\mu} + Z^{2,\mu}, \mu =1, \ldots, M\}
%% $$
%% with $m_1=0$,
%% $$
%% \Ccal_2:=\{Y^{2,\mu} + Z^{2,\mu}, \mu =1, \ldots, M\}.
%% $$
%% all the $(Y^{1,\mu}),(Y^{k,\mu})$ i.i.d. with
%% $$\P(Y^{1,1}_1=N)=\P(\tilde Y^{1,1}_1=1)=p=1-\P(Y^{1,1}_1=0)=1-\P(\tilde Y^{1,1}_1=0)$$
%% and $Z^\mu \in \{0,1\}^d$ i.i.d. for different $\mu$ are such that the coordinates are i.i.d. with a Ber(1/l)-distribution. The proofs of Propositions \ref{prop_pixel} - \ref{prop_segment} easily carry over to this new situation.

%% Another model which might be considered is to take
%% $\Ccal_1$ as above and
%% $$
%% \Ccal_2:=\{Y^{2,\mu} + Z^{2,\mu}, \mu =1, \ldots, M\}.
%% $$
%% with $Z^\mu \in \R^d$ are again i.i.d. and $\mathcal{N}_d(m, Id)$-distributed, where $Id$ is the $d  \times d$ identity matrix and
%% $$
%% m=(1,\underbrace{0, \ldots,0}_{l-1 \text{ times}}, 1,  \underbrace{0, \ldots,0}_{l-1 \text{ times}},1 \ldots 0).
%% $$
%% Also in this situation one checks that Propositions \ref{prop_pixel} - \ref{prop_segment} remain true ({\tt I haven't checked, but I assume we need some LDP/MDP estimates for Prop. \ref{prop_segment}}).
%% \fi

However, analyzing the data used in \cite{gripon_classification} one sees that our models above describe well the behavior of one class, but not that of two classes simultaneously. Indeed, there is some evidence, that the coordinates that take large values in a class are, also likely to take large values in another, but the variance is larger for those that take large values. To take this into account, we change our original model in the following way:
\begin{eqnarray*}
\Ccal_1:=\{N^\mu Y + m_1, \mu =1, \ldots, M\} \\\Ccal_2:=\{\tilde N^\mu Y + m_2, \mu =1, \ldots, M\}.
\end{eqnarray*}
%and
%$$
%\Ccal_2:=\{\tilde N^\mu Y + m_2, \mu =1, \ldots, M\}.
%$$
%% Again $m_1= (0, \ldots, 0)$ and $m_2=(1,\underbrace{0, \ldots,0}_{l-1 \text{ times}}, 1,  \underbrace{0, \ldots,0}_{l-1 \text{ times}},1 \ldots 0)$.
Moreover $Y$ is a vector of i.i.d. Bernoulli random variables with parameter $p$,  i.e.
$\P(Y_1=1)=p=1-\P(Y_1=0).$
Finally the $N^\mu$ and $\tilde N^\mu$ are i.i.d. random variables in $\R^d$ with positive, i.i.d. components, such that $\P[N^\mu_1\ge a]=1$, for some $a>0$.
Again we will take dictionaries $D_j$ that only contain one segment $w_j^1$ and $w_j^2$ of each class and we want to classify a word $w$ that is distributed as a word from $\Ccal_1$ correctly. Assume that
$w^1=Y N, w_2= m_2+Y \tilde N$ and $w=Y \overline{N}$.
Again we obtain
\begin{theorem}\label{thm2}
In the model described above assume that $d,l\to \infty$, $l \le d^{\frac 14}$, $p \ll \frac 1 l$, but $p \gg \max(\frac 1d,  \frac 1{a^2l})$,
then for $c=d$ and $c=1$ we have
$
\P(w \text{ is misclassified}) \to 1/2
$
while for $c= d/l$,  we have
$
\P(w \text{ is misclassified}) \to 0.
$
\end{theorem}
The proof is only a slight modification of the proof of Theorem \ref{thm1}.

\section{Experiments}

In this section we derive experiments on real-world data. In our experiments, we use three distinct DNNs. Two of them are related to vision tasks and perform feature extraction from raw input images, namely Inception V3~\cite{szegedy2015rethinking} and SqueezeNet~\cite{iandola2016squeezenet}. Both these networks have been trained using 1'000 classes from the ImageNet dataset. As far as Inception V3 is concerned, we use the features obtained before the first fully connected layer. It consists of a vector with 2'048 dimensions. The inputs are images scaled to 299x299 pixels. For the SqueezeNet network, we use the penultimate layer (containing 1'000 dimensions) as our feature extractor. Input images contain 227x227 pixels. The last DNN we use has been trained on AudioSet~\cite{gemmeke2017audio}, a dataset that consists of more than 2'000'000 distinct audio tracks extracted from videos on YouTube. The extracted features contain 1'280 dimensions which are the concatenation of ten 128 dimensions feature vectors, one per second of the corresponding audio track.

We perform tests on four datasets: CIFAR10, two subsets of ImageNet made of 10 classes sampled randomly from those that where not used to train the DNNs, and a subset of 10 classes used to train AudioSet.
\ifextended
The first one is CIFAR10. CIFAR10 is a set of tiny images made of 32x32 pixels belonging to 10 different classes. This very low resolution results in signals that are quite different from those used to train the DNNs. We then introduce two datasets extracted from ImageNet, named ImageNet1 and ImageNet2. They contain both 10 classes that were not used to train the DNNs. These signals are thus much more similar to those used to train the DNN than in the case of CIFAR10. Training sets contain 5'000 items per class for CIFAR10 and 1'000 items per class for ImageNet1 and ImageNet2. Test is performed on 10'000 items for CIFAR10 and about 2'200 for the two ImageNet subsets. For AudioSet, we consider classes that have been used to train the DNN. We have chosen 10 classes with similar number of elements in the dataset (radio, cat, hi-hat, helicopter, fireworks, stream, bark, baby/infant cry, snoring, train horn). The cardinality of training sets ranges from 2'000 to 5'000 elements, and we test on about 600 other elements. We removed elements belonging to multiple of these categories (AudioSet contains multilabels elements).
\else
More details are available in the extended version of this paper~\cite{extended}.
\fi

We use $k$-NN as our nonparametric method to obtain a classification accuracy. Note that in the case of $c$ segments, the decision is taken using $k c$ votes instead of $k$, since each subspace performs a $k$-NN. We observe that in all scenarios, the optimal solution corresponds to an intermediate number of segments. The case of AudioSet is interesting as there is a local maximum in accuracy which corresponds to 10 segments, which occurs when considering the ten 128 dimension feature vectors independently, but the global maximum is for 40 segments. Note that the complexity of the method does not depend on $c$, as both memory and number of operations boils down to the product of the number of training vectors and their dimension.
Table~\ref{experiments_table} summarizes our results.

\begin{table}\label{experiments}
  \begin{center}
    \small
    \begin{tabular}{|c|c|c|c|c|c|}
      \multicolumn{6}{c}{\textbf{Inception V3, 1-NN}}\\
      \hline
      $c$ & 1 & 4 & 16 & 64 & 256\\
      \hline
      CIFAR10 & 0.8519 & 0.8652 & \textbf{0.8781} & 0.8651 & 0.8347\\
      \hline
      ImageNet1 & 0.9328 & 0.9354 & 0.9424 & \textbf{0.9439} & 0.9081\\
      \hline
      ImageNet2 & 0.9438 & 0.9451 & \textbf{0.9524} & 0.9464 & 0.9171\\
      \hline
      \multicolumn{6}{c}{\textbf{Inception V3, 5-NN}}\\
      \hline
      $c$ & 1   & 4 & 16 & 64 & 256\\
      \hline
      CIFAR10   & 0.8689 & \textbf{0.8761} & 0.8759 & 0.8668 & 0.8461\\
      \hline
      ImageNet1 & 0.9389 & \textbf{0.9450} & 0.9429 & 0.9394 & 0.9202\\
      \hline
      ImageNet2 & 0.9467 & 0.9498 & \textbf{0.9511} & 0.9488 & 0.9303\\
      \hline
    \end{tabular}
    \begin{tabular}{|c|c|c|c|c|c|}
      \multicolumn{6}{c}{\textbf{SqueezeNet, 1-NN}}\\
      \hline
      $c$ & 1   & 5 & 20 & 100 & 200 \\
      \hline
      CIFAR10   & 0.6839 & 0.7069 & \textbf{0.7472} & 0.6890 & 0.6225 \\
      \hline
      ImageNet1 & 0.8854 & 0.8900 & \textbf{0.9001} & 0.8784 & 0.8466 \\
      \hline
      ImageNet2 & 0.8737 & 0.8802 & \textbf{0.8926} & 0.8669 & 0.8267 \\
      \hline
      \multicolumn{6}{c}{\textbf{SqueezeNet, 5-NN}}\\
      \hline
      $c$ & 1   & 5 & 20 & 100 & 200 \\
      \hline
      CIFAR10   & 0.7284 & 0.7483 & \textbf{0.7566} & 0.6954 & 0.6371 \\
      \hline
      ImageNet1 & 0.8985 & 0.8965 & \textbf{0.8980} & 0.8698 & 0.8501 \\
      \hline
      ImageNet2 & 0.8862 & \textbf{0.8901} & 0.8893 & 0.8591 & 0.8280 \\
      \hline
    \end{tabular}
    \begin{tabular}{|c|c|c|c|c|c|}
      \multicolumn{6}{c}{\textbf{AudioSet}}\\
      \hline
      $c$ & 1 & 2 & 10 & 40 & 160 \\
      \hline
      1-NN & 0.605 & 0.621 & 0.704 & \textbf{0.724} & 0.660 \\
      \hline
      5-NN & 0.564 & 0.649 & 0.704 & \textbf{0.727} & 0.668 \\
      \hline
    \end{tabular}

  \end{center}
  \caption{Accuracy of classification, depending on the feature extractor used, the dataset, the number of segments $c$ and the number of nearest neighbors $k$ the decision is based upon. Best scores are marked in bold.}
  \label{experiments_table}
\end{table}

\section{Conclusion}

Transfer learning is a popular method to obtain cutting edge descriptors that can be exploited to classify new data. When combined with nonparametric methods such as $k$-nearest neighbor search, it provides a lightweight incremental solution that is suitable for devices with limited energy or computational capabilities. We have shown that segmenting vectors to perform $k$-nearest neighbor search in obtained subspaces can result in significant improvements in accuracy. Moreover, this change has no cost on memory usage neither on the number of operations required to fulfill the task.
Interestingly, this method can be thought about as an alternative to increasing the number of neighbors $k$ to consider when taking a decision.

Future work include considering other downstream classification techniques such as support vector machines and logistic regression. 

\bibliographystyle{IEEEbib}
\bibliography{LiteraturDatenbank}

\end{document}